\documentclass[sn-chicago]{sn-jnl}

\jyear{2021}%

\theoremstyle{thmstyleone}%
\theoremstyle{thmstyletwo}%

\theoremstyle{thmstylethree}%

\usepackage{natbib}
\usepackage{diagbox}
\usepackage[center]{caption}

\begin{document}

\title[Learning Causal Mechanisms through Orthogonal Neural Networks]{Learning Causal Mechanisms through Orthogonal Neural Networks}


\author*[1]{\fnm{Peyman} \sur{Sheikholharam Mashhadi}}\email{peyman.mashhadi@hh.se}

\author[1]{\fnm{S\l{}awomir} \sur{Nowaczyk}}\email{slawomir.nowaczyk@hh.se}

\affil*[1]{\orgdiv{Center for Applied Intelligent Systems Research (CAISR)}, \orgname{Halmstad University}, \orgaddress{ \country{Sweden}}}


\abstract{A fundamental feature of human intelligence is the ability to infer high-level abstractions from low-level sensory data. An essential component of such inference is the ability to discover modularized generative mechanisms. 
Despite many efforts to use statistical learning and pattern recognition for finding disentangled factors, arguably human intelligence remains unmatched in this area. 

In this paper, we investigate a problem of learning, in a fully unsupervised manner, the inverse of a set of independent mechanisms from distorted data points.
We postulate, and justify this claim with experimental results, that an important weakness of existing machine learning solutions lies in the insufficiency of cross-module diversification. Addressing this crucial discrepancy between human and machine intelligence is an important challenge for pattern recognition systems.

To this end, our work proposes an unsupervised method that discovers and disentangles a set of independent mechanisms from unlabeled data, and learns how to invert them. A number of experts compete against each other for individual data points in an adversarial setting: one that best inverses the (unknown) generative mechanism is the winner. We demonstrate that introducing an orthogonalization layer into the expert architectures enforces additional diversity in the outputs, leading to significantly better separability. Moreover, we propose a procedure for relocating data points between experts to further prevent any one from claiming multiple mechanisms. We experimentally illustrate that these techniques allow discovery and modularization of much less pronounced transformations, in addition to considerably faster convergence.}

\keywords{Causal Mechanism, Orthogonalization, Diversity, Unsupervised Learning, Ensemble learning}



\maketitle

\section{Introduction}\label{sec1}

Understanding cause-effect relations and the mechanisms by which they are related is a primary element of human intelligence. A key aspect related to such mechanisms is modularity \citep{scholkopf2019causality}. Intuitively, the generating processes of different abstract concepts can be analyzed and modeled separately, or with minimal interactions.
 
Equipped with ways of discovering such mechanisms, humans are able to use them as building blocks in countless combinations \citep{vankov2020training}. As a narrow manifestation of this concept, Humboldt referred to a language as a system that ``makes infinite use of finite means,'' later quoted by Noam Chomsky many times. Since an infinite combinations of sentences can be generated from a finite number of grammatical rules, arguably the understanding of core mechanisms is the key component towards broader generalization. 

As stated by \citep{chollet2019measure}, the ability to generalize is a spectrum, with symbolic AI offering no generalization, rising up towards local generalization provided by machine learning (specialist AI), and with the ultimate goal of Artificial General Intelligence.
Stepping into the new era of deep learning with hierarchical knowledge representations has the potential to offer the ability to adapt to new situations far from the learned manifold of seen data. We believe that robust and fast methods for identifying independent generative mechanisms, or transformations, directly from the data is an important next step for the AI field, especially as we move towards ``knowledge creation'' \cite{Nowaczyk2020}.

As a special case, finding invariant features has been one of the prime concerns of pattern recognition systems \citep{goodfellow2009measuring}. Needless to say, the promise is to recognize patterns of interest under different transformations. For instance, a Convolutional Neural Network (CNN), presented with enough relevant samples during the training, can learn characteristics of classes of interest through a set of functions mapping original features into a manifold where the effects of different mechanisms manifest in similar ways \citep{dhillon2020convolutional}.
 
We argue that despite the tremendous success of the deep learning field, they lack the true generalization perspective \citep{jakubovitz2019generalization,baldock2021deep}. The first issue is that the transformations learned within a trained model are entangled \citep{xie2020explainable}. This 
lack of modularity, as we discussed earlier, makes it impossible to come up with new, ``creative'' combinations of learned concepts \citep{battaglia2018relational}. Finding disentangled factors is substantially different from building mechanisms that bring back data points of a given category to a common region of space. More concretely, consider an image dataset containing faces that are either rotated or stretched. A new image, on which both rotation and stretching are applied, will lie far from the manifold learned of such data. Nevertheless, models that captured rotation and stretching mechanisms separately, can bring this data point back to the manifold, and subsequently recognize.

Our research extends the study of \citet{parascandolo2018learning}, which proposed several experts to adversarially compete against each other for transforming data points. The goal for each expert is to find the inverse of a particular mechanism, and thus fool a discriminator.

We follow the same setting, providing two substantial contributions over the previous work: the ability to invert less pronounced transformation, and greatly reduced convergence time.  We achieve these by proposing, first, to augment experts with an orthogonalization layer, thus forcing their respective outputs to be orthogonal to each other. This leads to higher diversity among experts, which causes the discriminator to provide a more informative signal. Secondly, we introduce data point relocation between experts, discouraging any one of them from claiming multiple transformations. The key finding in this work is the discovery of a previously unknown bottleneck in terms of separability power, namely, that the discriminator part of the adversarial architecture is an insufficient diversification factor. By adapting a recently-proposed orthogonality layer (originally intended for ensemble learning), our model gains the ability to invert less pronounced transformations. This is a fundamental difference that could not be replicated in the existing method simply by using more computations, neither in terms of time nor amount of data.

The remainder of the paper organized as follows. Section \ref{related-works} describes related work; Section \ref{problem-setting} presents problem formulation; Section \ref{methodology} explains the proposed method, and discusses key differences from state of the art; Section \ref{experiments} contains results of our experiments. Finally, Section \ref{conclusion} concludes the paper.

\section{Related work}
\label{related-works}

The most relevant research that we draw from is \cite{parascandolo2018learning}. We share the same goal, i.e., to find inverse of a set of mechanisms in an unsupervised fashion. Examples are available of both untouched data and distorted data; however, there is no information about mechanism, or transformation, was applied to a particular data point. The key idea is for a set of experts to compete against each other for data points, where the one best capable of reversing the mechanism wins and gets trained further. 

Along similar lines, \cite{von2020towards} proposed decomposition of complex visual scenes using ensemble of experts competing for different parts of a visual scene. An important assumption in their study is that a scene is layered composition of depth-ordered objects. In order for each expert to become specialized on a different part, firstly, they are equipped with an attention network. Then, the attended region is mapped into a latent space and, through a decoder, reconstructs the unoccluded object shape. 

\citet{locatello2018competitive}
proposed a clustering procedure adopting generative models. Each data point is supposed to be generated by one causal mechanism. Generators (chosen to be VAEs) compete against each other with the purpose of generating realistic data points. The method is the generative counterpart of k-means, since an assignment function assigns data points to different partitions (equal to the number of generators). The generators compete to win over the distribution formed by each partition. 

\citet{fox2019advocacy} proposed a method called ``advocacy learning'' for a supervised setting, where a set of advocates (decoders) generate class-conditional representations to encourage a judge (classification model) that an input data point belongs to their class. The output of each advocate forms an attention map that, after modulation with the input, will be fed to the judge. The idea is that each advocate learns an attention map which eases the judge's classification task.

Clearly, the performance of all these methods greatly depends on the discriminator providing the right signal, i.e., assigning distorted data points to the ``correct'' expert. This can only be done reliably given enough diversity among the experts. In an extreme case, if all the experts generate equivalent output, the discriminator could not reliably differentiate between the mechanisms. We claim that existing competitive learning approaches are not sufficient, and thus our key contribution is adapting an enforced diversity module that \cite{mashhadi2021parallel} proposed for ensemble learning into the learning causal mechanisms setting. Once the diversity is enhanced by making the outputs of all experts orthogonalized to each other, the crucial pressure on the discriminator is lowered significantly.

\section{Problem Setting}
\label{problem-setting}

Let $\mathcal{P}$ be the empirical distribution of original images (before transformation) defined on pixel space $\mathbb{R}^d$. Also, let's assume that there are $N$ different transformations (mechanisms) denoted by $M_1, M_2,..., M_N$ that can be applied on the original images. Let's denote the resultant empirical distribution of applying the mechanisms by $Q_1, Q_2, ..., Q_N$ where $Q_i = M_i(\mathcal{P})$. 

For training, the model is provided with two separate datasets: $D_P$, with data points from the original distribution $\mathcal{P}$, and $D_Q$ with data points sampled  i.i.d from transformed distributions $Q_1, Q_2, ..., Q_N$. It is important to notice that nothing is indicating which transformed data point in $D_Q$ corresponds to which original data point in $D_\mathcal{P}$. If that was the case, one could feed a model with original data points as input and the transformed data points from the same transformation class as output. 
Alas, in real world, the supervised scenario is rare compared to the unsupervised ones. 

For simplicity and interpretability of results, let us assume that the number of experts is equal to the number of mechanisms, denoted by $E_1, E_2, ..., E_N$. The desired outcome is that each expert becomes specialized in inverting a single transformation. The original paper \cite{parascandolo2018learning} showcases natural relaxations of this scenario, which we omit due to space constraints.

\section{Methodology}
\label{methodology}

In this section, we first explain the overall adversarial process of learning mechanisms, followed by the proposed model architecture and its motivation.

\subsection{Learning mechanisms via competing experts}
\label{subsec:competetive learning}

As proposed by \citep{parascandolo2018learning}, the adversarial process of learning mechanisms is based on making experts compete over data points. For every transformed data point $x'$ from $D_Q$ in  $\mathbb{R}^d$, each expert generates a new data point in $\mathbb{R}^d$, trying to trick the discriminator into recognizing their generated images as coming from $\mathcal{P}$. In other words, each expert $E_i$ tries to maximize the output of the discriminator denoted by $c_i$, where ($c_i:\mathbb{R}^d \rightarrow \mathbb{R}$).  

In the training process, the corresponding discriminator's outputs for all the experts $c_i = c(E_{i}(x'))$ are compared to each other. Then, the parameters $\theta^*$ of only the winning expert $E^*$, the one with the highest $c_i$, are updated. This means that the one best able to inverse a specific mechanism (manifested in $x'$), will be reinforced to improve even further.

Along with updating the winning expert, the discriminator is also updated, however, it is taking into account the outputs of all the experts. Quite similar to the standard GAN's discriminator loss function \citep{goodfellow2020generative}, the following is used to update the discriminator.

\begin{multline}
\label{disctiminator-loss}
\max_{\theta_D} \quad \mathbb{E}_{x \sim \mathcal{P}} \; \log\Big(D_{\theta_D}(x)\Big) +  \frac{1}{N^{'}}\, \sum_{i=1}^{N^{\prime}} \; \mathbb{E}_{x^{\prime} \sim Q} \Big(\log(1 - D_{\theta_{Q_D}}(\mathbb{E}_{\theta_i)}(x^{\prime})\Big)
\end{multline}

The difference between Eq.~\ref{disctiminator-loss} and the default GAN's discriminator loss function is that in our case the second term is the summation over all the experts (generators). The winning expert loss function is exactly the standard GAN's generator loss:

\begin{equation}
    \label{gramschmidt}
    \min{\theta_{E^*}} \quad \mathbb{E}_{{x^{\prime}} \sim Q} \; 1 - \log\Big(D_{\theta_D}(E^*(x^\prime))\Big)     
\end{equation}

\subsection{Proposed architecture}

\begin{figure}[b]
  \centering
  \includegraphics[scale=.4]{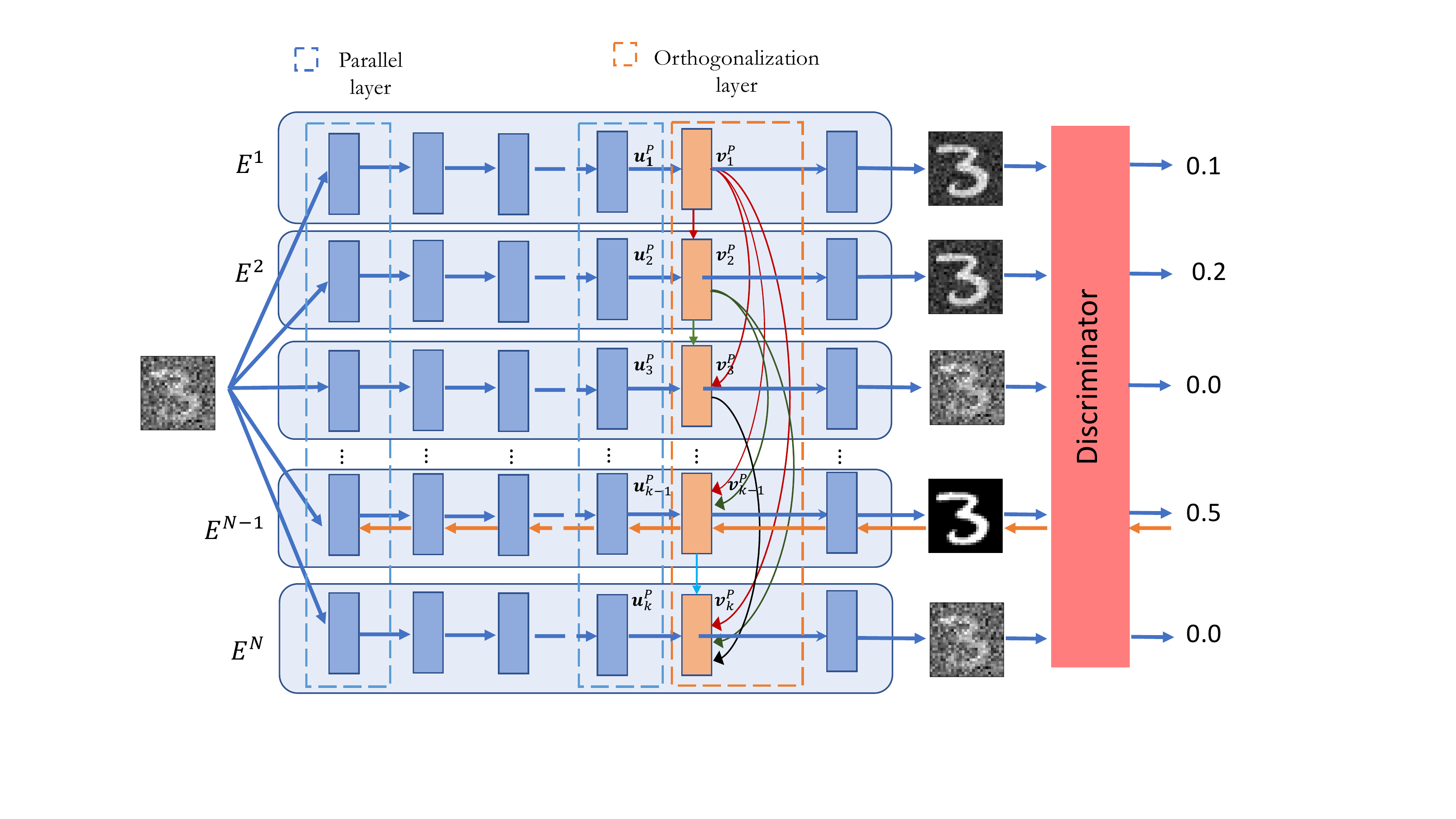}
  \caption{Parallel Orthogonal Deep Neural Network (PODNN)}
  \label{fig:podnnMechanism}
\end{figure}

We propose the following architecture to learn the inverse of mechanisms. As seen from Figure \ref{fig:podnnMechanism}, first a number of parallel experts are created. The main novelty of our architecture is adding the ``Orthogonalization Layer.'' Following \cite{mashhadi2021parallel}, the role of this layer is restricting the output of experts at a specific location to become orthogonal to each other. The motivation is to enforce diversity, and we demonstrate in the next section that this increased diversity translates into higher effectiveness of the discriminator.

The most crucial challenge in this competitive learning schema is providing effective discriminating signal to correctly assign experts to mechanisms. The better signal the discriminator provides, the more effective and faster the training of the experts -- every mistake leads to confusing an expert by training it on an inappropriate mechanism. Without the orthogonalization layer, the only way to diversify the experts is through discriminator-provided training signal, as described in Section \ref{subsec:competetive learning}. Although this process is very promising, we argue that the learning becomes significantly more efficient if an explicit diversity technique is incorporated.

In the proposed architecture, we enforce expert diversity using the orthogonalization layer. 
The output of the first expert at layer $P$, $\mathbf{u}_1^P$, is untouched ($\mathbf{v}_1^P$); then, the output of the second expert, $\mathbf{u}_2^P$, is orthogonalized to the $\mathbf{v}_1^P$ using the Gram-Schmidt orthogonalization, see Equation \ref{eq:gramschmidt}. The result of this orthogonalization is denoted by $\mathbf{v}_2^P$ in Fig. \ref{fig:podnnMechanism}. Then, the output of the third expert, $\mathbf{u}_3^P$, is orthogonalized to the previous two experts, resulting in $\mathbf{v}_3^P$. This process continues to the last expert. 
More detailed description of the details on diversity and error backpropagation through the orthogonalization layer is included in \citep{mashhadi2021parallel}.

\begin{equation}
\label{eq:gramschmidt}
    \mathbf{v}_k = \mathbf{u}_k - \sum_{i=1}^{k-1} \dfrac{(\mathbf{v}_i,\mathbf{u}_k)}{(\mathbf{v}_i, \mathbf{v}_i)} \mathbf{v}_i
\end{equation}

An important distinction between our work and the earlier study by \citet{parascandolo2018learning} is that in the latter, the discussion is made around learning independent mechanisms. Both mechanisms and experts are independent in the sense that they do not inform or influence each other. However, in our architecture, the experts do interact and exchange information, throughout the orthogonalization layer. \citet{besserve2018counterfactuals}
introduced the need for going beyond statistical independence to disentangle latent factors. They argued that many interesting transformations to latent factors are actually statistically dependent and proposed a non-statistical framework based on counterfactual manipulation to find modular structures. We take this transition to yet another level, arguing that there are advantages for modular mechanisms to influence and inform each other. Thinking Bayesian, if the new evidence alters the posterior of what a mechanism is, the updated mechanism should resonate with other mechanisms, and if it does not, it should influence the other mechanisms. For example, if one's concept of what ``right translation'' means changes, the ``left translation'' (and other translations) might need to be adjusted accordingly.  

\subsection{Data points relocation mechanism}

While the orthogonalization layer leads to higher diversity of experts and better separability of mechanisms among them, still, in some cases the confusion between experts persists. To further discourage an expert from claiming multiple transformations, we propose the data points relocation mechanism.

The main idea behind the relocation mechanism is to separate a percentage of low-confidence data points from an expert and assign them to expert which has not won any data point. The level of confidence, or proficiency, of an expert on any given data point is measured by the discriminator score. The lower be the discriminator score, the smaller the confidence. However, selecting which expert should ``give up'' a data point is challenging, since there is no way to know that an expert has won more than one transformation (due to the unsupervised nature). We propose to use the standard deviation of the euclidean distances of discriminator's last hidden layer on the data points claimed by an expert as a proxy of ``ambition level'' of the expert. We argue that experts with higher standard deviation are most likely the experts that are trying to claim more than one transformation. The motivation behind this idea is that it would be difficult for an expert to find a hidden space in which data points from different transforms lie very close to each other; in comparison, for an expert winning only one transformation, this task is significantly easier, and the resulting space is more compact.   

Finally, since the above modifications lead to significantly faster convergence of experts compared to previous work, we argue that the competitive learning approach is only needed in the beginning. After all the experts converge, there is no need anymore to continue the training through the discriminator. Therefore, from that point forward, we only use the discriminator to assign the data points to the specialized experts, while training then standalone. This saves computation, since the error does not need to get propagated from the discriminator to all experts. 
The pseudocode of the proposed method is depicted in Algorithm \ref{alg:algorithm1} and the data points relocation mechanism is depicted in Algorithm \ref{alg:algorithm2} \footnote{source code available at: \url{https://github.com/causalPODNN/causalpodnn}}.

\begin{center}
\begin{algorithm}[t!]
\caption{Learning causal mechanisms through PODNN }
\label{alg:algorithm1}
\textbf{Input}: PODNN containing $N'$ experts; $D$: discriminator 
$X_\mathcal{P}$: data sampled from $\mathcal{P}$; $X_Q$: data sampled from $Q$ \\
\begin{algorithmic}[1] 
\State $t=1$
\While{TRUE}

\State $\triangleright$ Sample mini-batches
\State $x_\mathcal{P},x_Q \gets sample(X_\mathcal{P}),sample(X_Q)$.
\State $\triangleright$ Score all the experts' output using $D$ and calculate output of the
discriminator's last hidden layer on all data point in

\For{$i=1$ to $N'$}
\State $c_i \gets D(E_i(x_Q))$
\State $h_i \gets D[\text{last hidden layer}](E_i(x_Q))$
\EndFor
\State $\triangleright$ Update discriminator

\State
$\displaystyle \theta_{D}^{t+1} \gets   \max_{\theta_{D}^{t}} \left(\nabla \mathbb{E}_{x_\mathcal{P}} \log D(x_\mathcal{P}) + \nabla\left(\frac{1}{N'}\right) \sum^{N'}_{i=1}  \log(1-\mathbb{E} (c_i))\right)$

\State $dp_1,dp_2,...dp_{N'} \gets$ assign data points to each expert according to $c\}_{i=1}^{N'}$ values 

\State $\triangleright$ Relocation mechanism
\If {exists at least one expert without any claimed data point}
\State $idx_L \gets$ select a ramdon expert among expert without any claim 
\State $ dp_1,dp_2,...,dp_{N'} \gets$ \\
\hfill{relocation\_algorithm $\left(h\}_{i=1}^{N'},c\}_{i=1}^{N'},dp\}_{i=1}^{N'},idx_L\right)$}
\EndIf

\State $\triangleright$ Update all experts
\For{$i=1$ to $N'$}
\State 
{$\displaystyle \theta_{E_i}^{t+1} \gets \max_{\theta_{E_i}^{t}} \left(\nabla \mathbb{E}_{dp_i} \log D(c_i) \right)$}

\EndFor

\If{convergence}
\State \textbf{break}
\EndIf
\State $t \gets t+1$
\EndWhile
\State $\triangleright$ Stop training discriminator and utilize it only for scoring the experts
\State
Continue training experts on the data points they are claiming
\end{algorithmic}
\end{algorithm}
\end{center}


\begin{algorithm}[tb]
\caption{Data point relocation algorithm}
\label{alg:algorithm2}
\textbf{Input}: 
$h_1,h_2,...,h_{N'}$: outputs of the discriminator's last hidden layer for data points claimed by each expert\\
$c_1,c_2,...,c_{N'}$: score of all each experts’ output using D on all data point in $x_Q$\\
$dp_1,dp_2,...,dp_{N'}$: data points assigned to each expert \\
$idx_L$: index of an expert without any claim
\\
\textbf{Parameter}: 
$rp$: relocation percentage, percentage of low-confidence data points to be relocated

\begin{algorithmic}[1] 
\For{$i=1$ to $N'$}
\State $dist_i \gets$ pairwise euclidean distance within $h_i$
\State $std_i \gets$ standard deviation of $dist_i$
\EndFor

\State $\triangleright$ Find the expert with highest std
\State $idx_W \gets \mathrm{argmax}(std_1,std_2,...,std_{N'})$

\State $\triangleright$ Select low-confidence data points claimed by expert\\ \hspace{0.2cm} $idx_W$ 

\State $low\_confidence\_dp \gets$ select $rp\%$ of data points sorted \\\hspace{3.6cm}in ascending order of $c_{idx_W}$

\State remove $low\_confidence\_dp$ from $dp_{idx_W}$ and to $dp_{idx_L}$

\State \textbf{return} $dp_1,dp_2,...,dp_{N'}$

\end{algorithmic}
\end{algorithm}

\section{Experiments}
\label{experiments}

We run several experiments using MNIST and Fashion-MNIST datasets. Three different categories of transformations are used: translation, adding noise, and contrast inversion. For the translation, in different experiments, depending on the desired number of mechanisms, multiple directions of translation are used: right, left, up, down, right-up, right-down, left-up, and left-down. The goal of the model is identifying, or separating, these transformations (mechanisms), and then learning how to reverse each one.

We conduct three sets of experiments, each with a specific purposes. The first experiment demonstrates the fast convergence of the proposed architecture, in comparison to the architecture without orthogonalization. The second experiment focuses on the different roles of the experts and the discriminator, and shows that it is the discriminator that benefits the most from the orthogonalization. Moreover, it highlights the drawback of over-reliance on the discriminator, especially for less pronounced transformations -- where our method,
due to enforced diversity, solves problems than the original counterpart could not handle.
The third experiment justifies the relocation technique, by showcasing how it prevents an expert from claiming multiple mechanisms.

\subsection{Faster convergence}
\label{sec:convergence}

\begin{figure*}[hbt!]
  \centering
  \includegraphics[width=\linewidth]{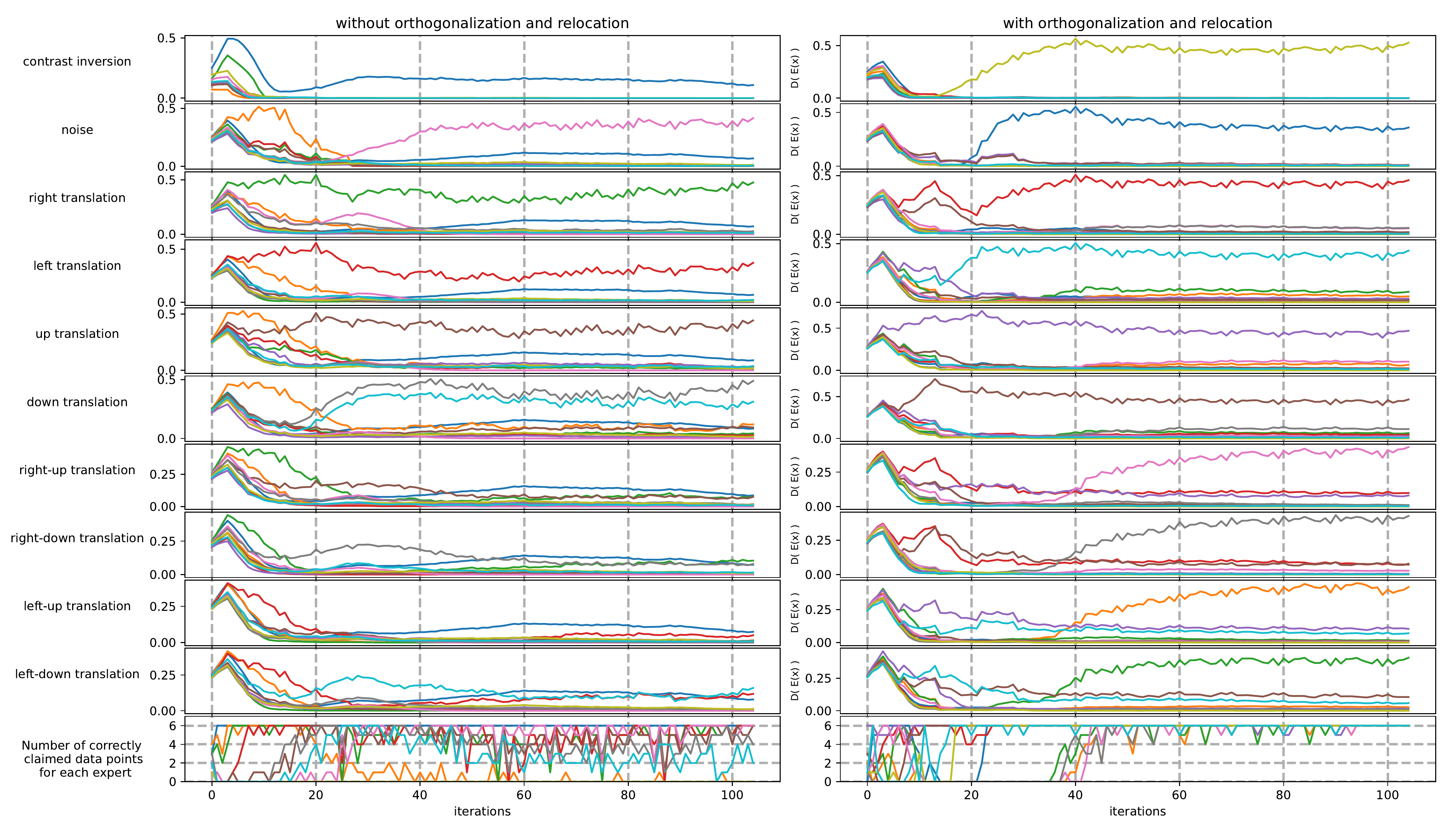}
  \caption{ MNIST- Comparison of experts' performance without orthogonalization (and relocation) and with orthogonalization (and relocation). Each line color represents one expert, and each row corresponds to one transformation. Experts' performance is measured by discriminator scores.}
  \label{fig:convergence_speed_both}
\end{figure*}

\begin{figure}[h]
  \centering
  \includegraphics[width=\linewidth]{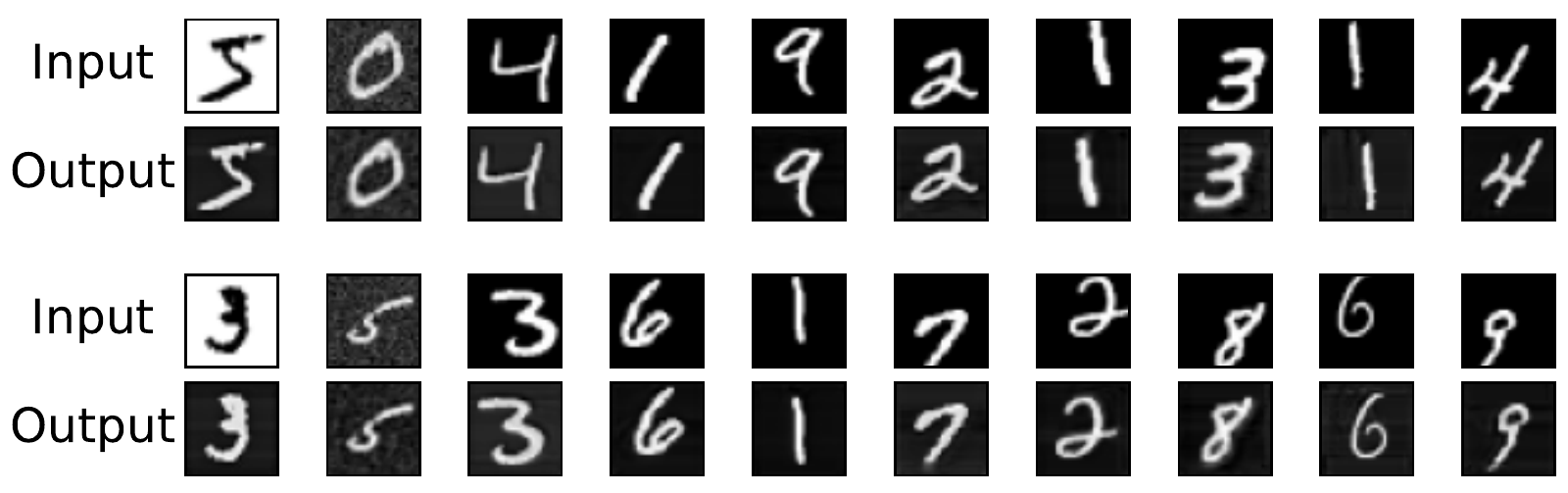}
  \caption{MNIST- In each group of rows, the top row shows the input to the experts, and the button row shows the output of an expert with the highest discriminator score}
  \label{fig:digits}
\end{figure}

\begin{figure}[h]
  \centering
  \includegraphics[width=\linewidth]{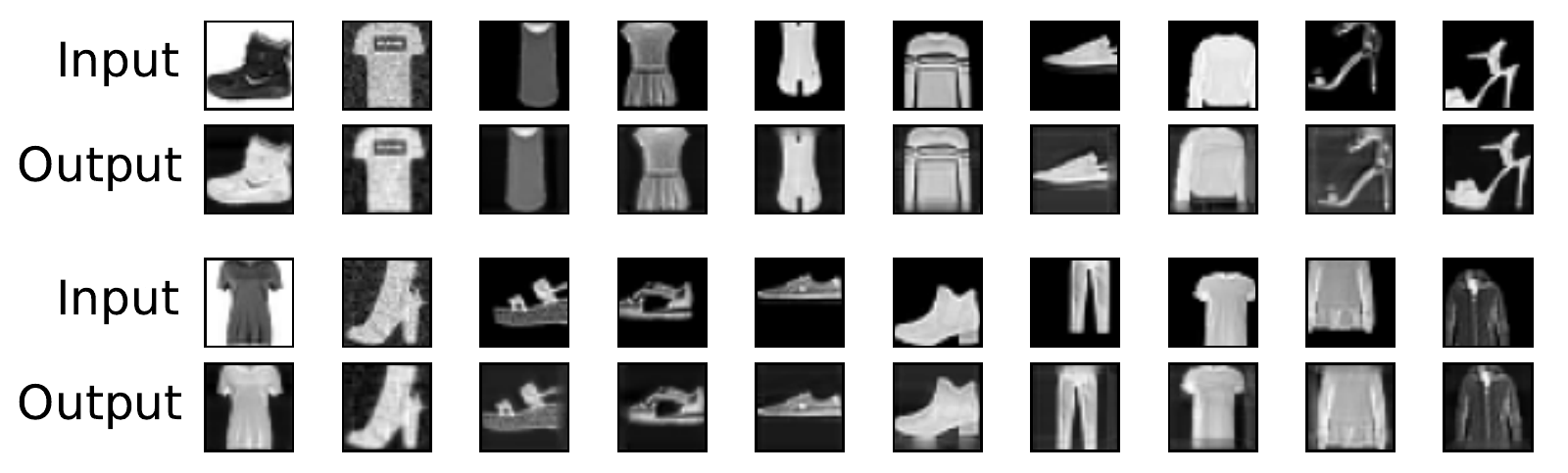}
  \caption{Fasion-MNIST- In each group of rows, the top row shows the input to the experts, and the button row shows the output of an expert with the highest discriminator score}
  \label{fig:fashion}
\end{figure}

\begin{figure*}[hbt!]
  \centering
  \includegraphics[width=\linewidth]{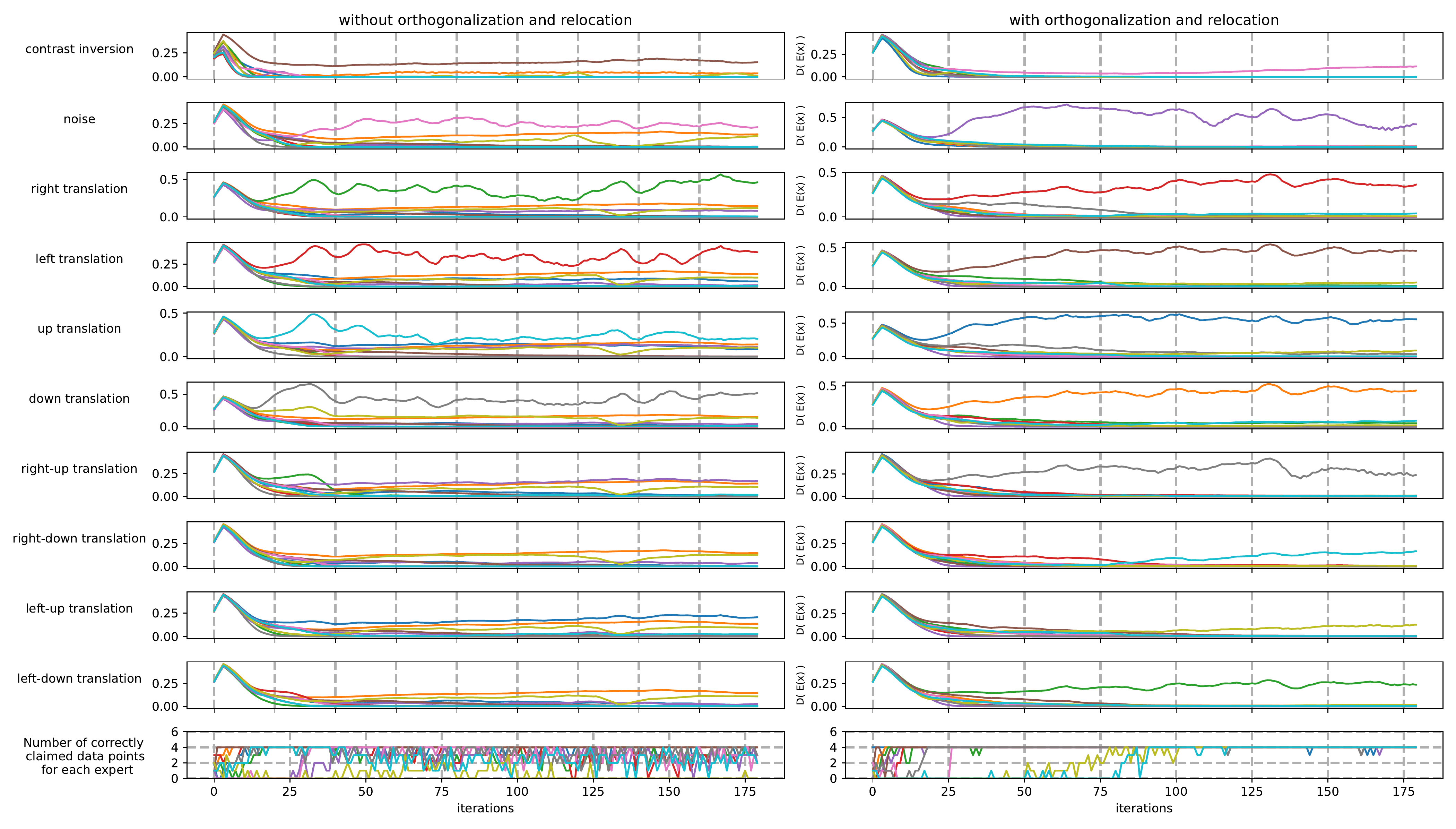}
  \caption{Fasion-MNIST- Comparison of experts' performance without orthogonalization (and relocation) and with orthogonalization (and relocation). Each line color represents one expert, and each row corresponds to one transformation. Experts' performance is measured by discriminator scores.}
  \label{fig:convergence_speed_both_fashion}
\end{figure*}

Figures \ref{fig:convergence_speed_both} and \ref{fig:convergence_speed_both_fashion} compare the convergence speed of the proposed method with its counterpart, on MNIST and Fashion-MNIST datasets, respectively. The left column is without orthogonalization layer and data point relocation, while the right column is the proposed method with orthogonalization and data point relocation mechanism. The last row in the figure shows the number of ``correct'' data points claimed by each expert, i.e., coming from the transformation they are most specialized in. For the first 10 rows, the vertical axis shows the discriminator score of experts. For the left column, after more than 100 iterations, there is still confusion among experts, while for the right column, after iteration 40, each experts is on the right track -- claiming only the points from a single transformation, different from all the others. Figures \ref{fig:digits} and \ref{fig:fashion} illustrate the result of trained experts on random subsets of transformed images.  

In order to showcase the consistency of the convergence, we define ``convergence iteration'' as the iteration after which every expert wins the majority of their data points from exactly one transformation. This definition is equivalent, and more intuitively understood, as the last swap in visualization such as the graphs of Figure \ref{fig:convergence_speed_both} (MNIST) and \ref{fig:convergence_speed_both_fashion} (Fashion-MNIST).

\begin{figure}[h]
  \centering
  \includegraphics[scale=0.14]{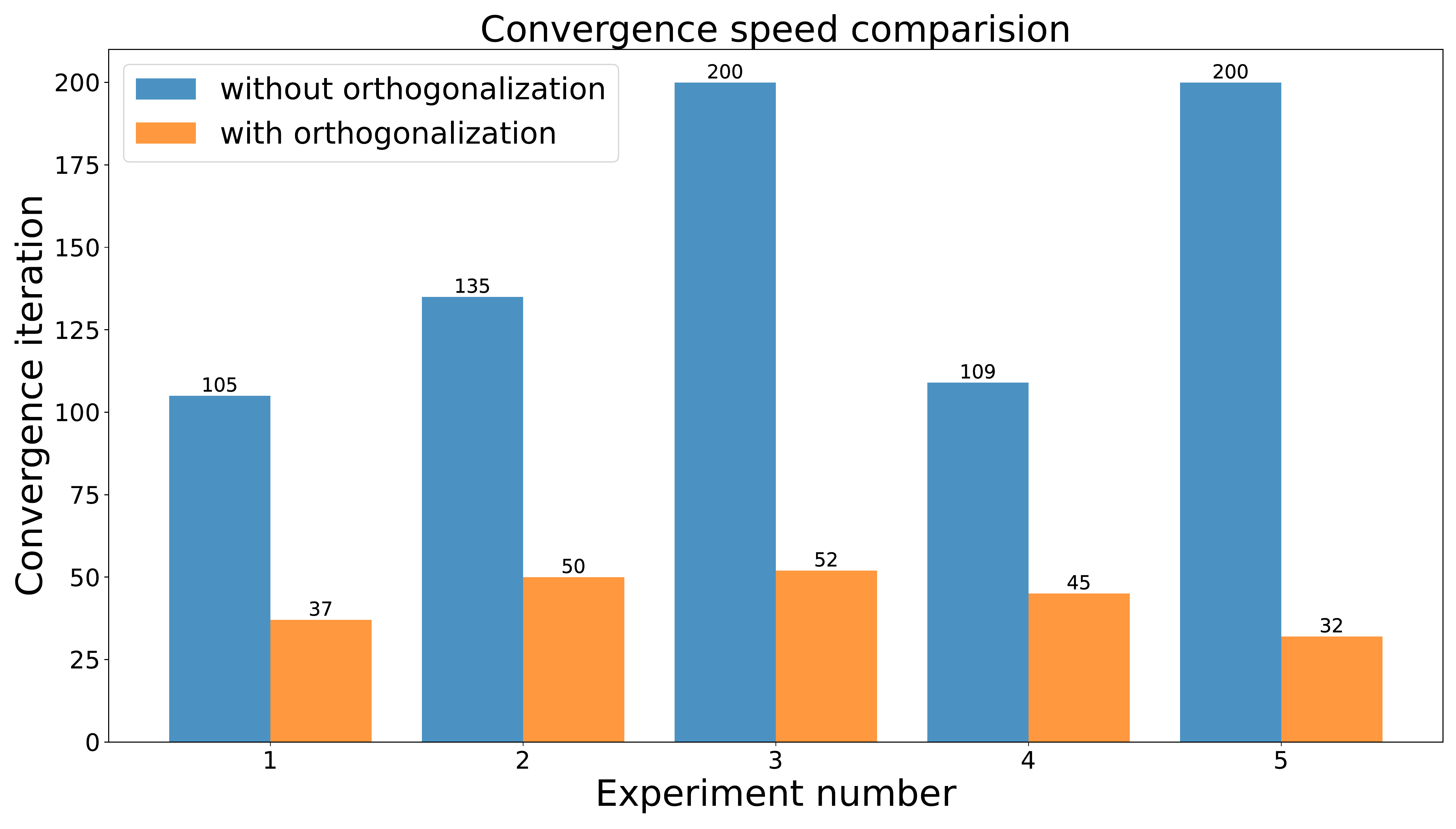}
  \caption{Convergence speed comparison between ``without orthogonalization'' (and relocation) and ``with orthogonalization'' (and relocation) on 5 separate runs}
  \label{fig:convergence_speed}
\end{figure}

In Figure \ref{fig:convergence_speed_both} the right column, the last swap occurred around iteration 40 (in the seventh row), while in the left column, convergence has not happened yet. Comparison of the convergence iteration of the proposed method with its counterpart across five different runs is shown in Figure \ref{fig:convergence_speed}. The experiment was limited to 200 iterations, so in the third and fifth runs the experts did not converge at all. 
It can be seen that the orthogonalization leads to drastic improvement. Moreover, the variance of the convergence iteration is also much lower for the proposed method. In Figure {\ref{fig:convergence_speed_both_fashion}}, the improvement is even more notable as in the right column, the last swap occurred at iteration 85. In contrast, for the left column, even after 180 iterations, the orange expert prevailed three tasks. It is observed that for Fashion-MNIST, without organization and data point relocation, for different runs convergence never happened. Figure {\ref{fig:fashion}} illustrates the result of trained experts with orthogonalization and data point relocation on a random subset of transformed images.

The rest of the experiments are reported only for  MNIST dataset as our proposed method's improvement is not as strong as it is in the case of Fashion-MNIST. In other words, we chose the one that is less favorable to our method, as a more fair comparison. 

\subsection{Time complexity}

The computation time associated with the addition of the orthogonalization layer is noticeable, but way lower than the gains provided. In general, the proposed architecture takes on average 10\% more time to train than the one without orthogonalization. However, since our convergence iteration is as shown in Figure \ref{fig:convergence_speed} is on average 3.5 times faster than its counterpart, the orthogonalization cost is negligible. In addition, since after experts' convergence we stop training them through the discriminator and continue training them standalone, we save additional computations by removing the discriminator from the backpropagation.

\subsection{Diversity helps the discriminator}

Given that orthogonalization promotes diversity among experts, we argue that the emerged diversity is particularly helpful for the discriminator, allowing it to provide more effective discriminating signal. Looking at the system (experts and discriminator) as a whole, it is not straightforward to understand which part benefits most from the embedded diversity component. In other words, we need to disentangle the discriminator from the experts to analyze the effect of diversity on them. 

We intend to do it by varying the severity of transformations. First, we replace the discriminator with an oracle, meaning that the discriminator knows in advance what transformation is applied on any image $x'$. We then consider different severity of the transformations, for example, the translation distance. We demonstrate that the experts are mostly invariant to that. For instance, translating four pixels to the right would not make it much more difficult for the experts than translating one pixel to the right.

To evaluate the experts' performance in undoing the transformations, their outputs are fed to a CNN trained on MNIST. For this experiment, the transformations are chosen to be various translations, with different severity, measured in pixels. Moreover, the relocation mechanism is factored out for this experiment to focus on the orthogonalization aspect as our main driving force of diversification. Table \ref{tab:faked_discriminator} reports the results, in terms of accuracy, without and with orthogonalization. First, it can be seen that as we increase the severity from 1px to 5px, the accuracy of experts remains essentially the same -- there is no difference in difficulty. Moreover, regardless of the transformation severity, the difference in accuracy of experts with and without orthogonalization is negligible.

\begin{table}[b]
\begin{center}
\caption{Effect of transformation severity on experts' performance (in terms of accuracy of MNIST-trained CNN.}
\label{tab:faked_discriminator}
\scriptsize
\centering
\begin{tabular}{|l|*{5}{c|}}
\hline
\backslashbox{Method}{Severity}
& 1px & 2px & 3px
& 4px & 5px \\\hline\hline
Without Orth & 0.9720 & 0.9720 & 0.9718 & 0.9706 & 0.9687\\\hline
With Orth & 0.9704 & 0.9702 & 0.9682 & 0.9652 & 0.9633 \\
\hline
\end{tabular}

\end{center}

\end{table}

This indicates that the key to improved performance in Section~\ref{sec:convergence} lies in the benefits that the diversity brings to the discriminator. It is intuitively understandable why the discriminator would be very susceptible to the severity of transformations. When they are very minor, such as translation by one pixel, it is difficult to provide an effective discriminating signal. The outputs from all the experts look as ``reasonable'' candidates belonging to $\mathcal{P}$. Identifying the ``winning'' expert is much easier for more pronounced transformations, since the outputs of different experts are more diverse.

Therefore, we repeat the previous experiment with transformations of difference severity using a ``real'' discriminator. Our goal is to investigate whether orthogonalization leads to a better discriminating signal
to the competing experts. We perform this experiment on two different severities of transformations: 2px and 5px. We expect that in the latter case the discriminator would not have much difficulty providing an effective discriminating signal to the experts, both in the setting with and without orthogonalization. On the other hand, when the transformations are less pronounced, the discriminator will work much better in case of the experts with orthogonalization than without it.

\begin{figure}[t]
  \centering
  \includegraphics[width=\linewidth]{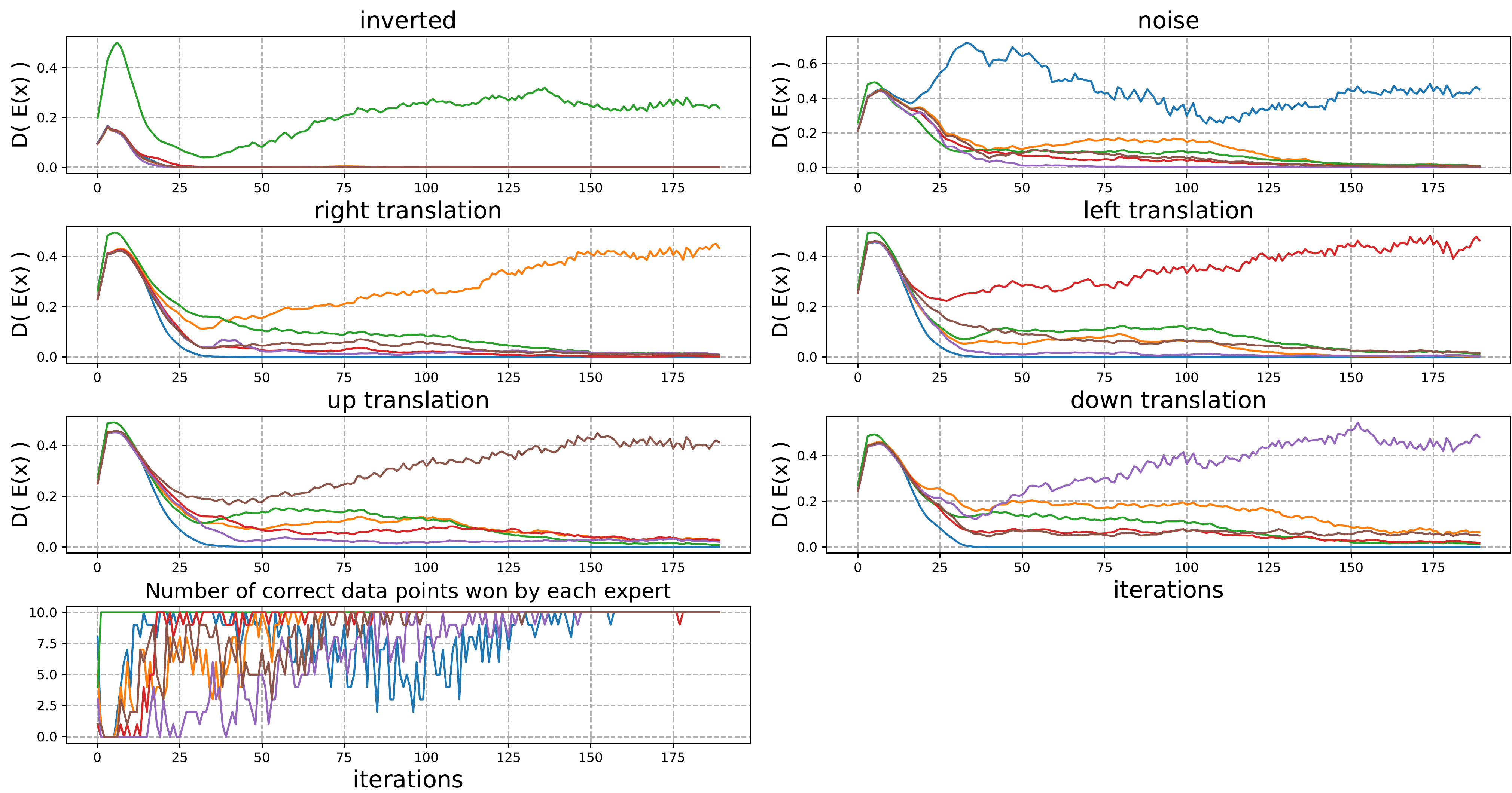}
  \caption{Without orthogonalization, transform severity 5px}
  \label{fig:severity_notorth_5px}
\end{figure}

\begin{figure}[t]
  \centering
  \includegraphics[width=\linewidth]{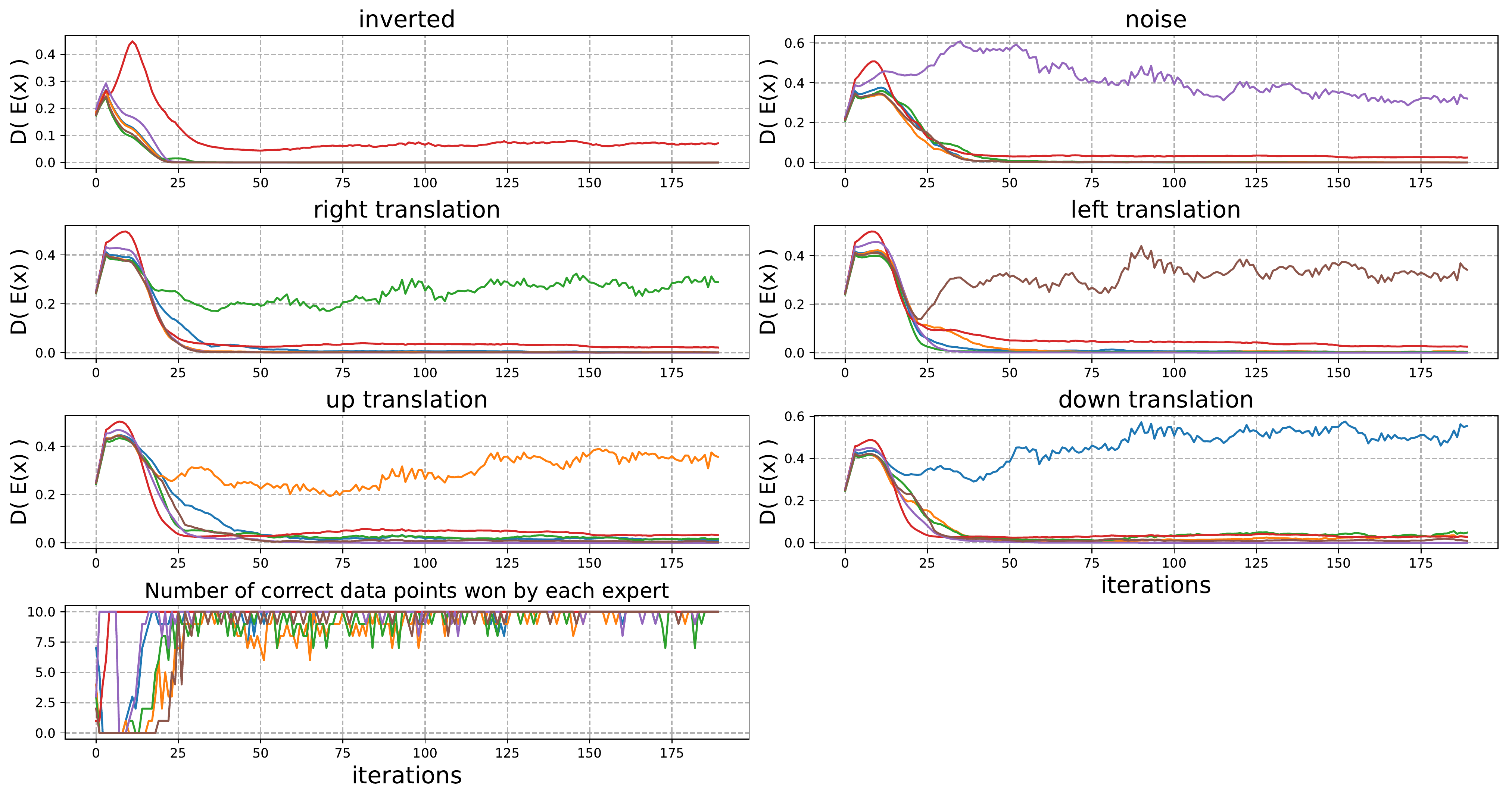}
  \caption{With orthogonalization, transform severity 5px}
  \label{fig:severity_orth_5px}
\end{figure}

\begin{center}
\begin{figure}[t]
  \centering
  \includegraphics[width=\linewidth]{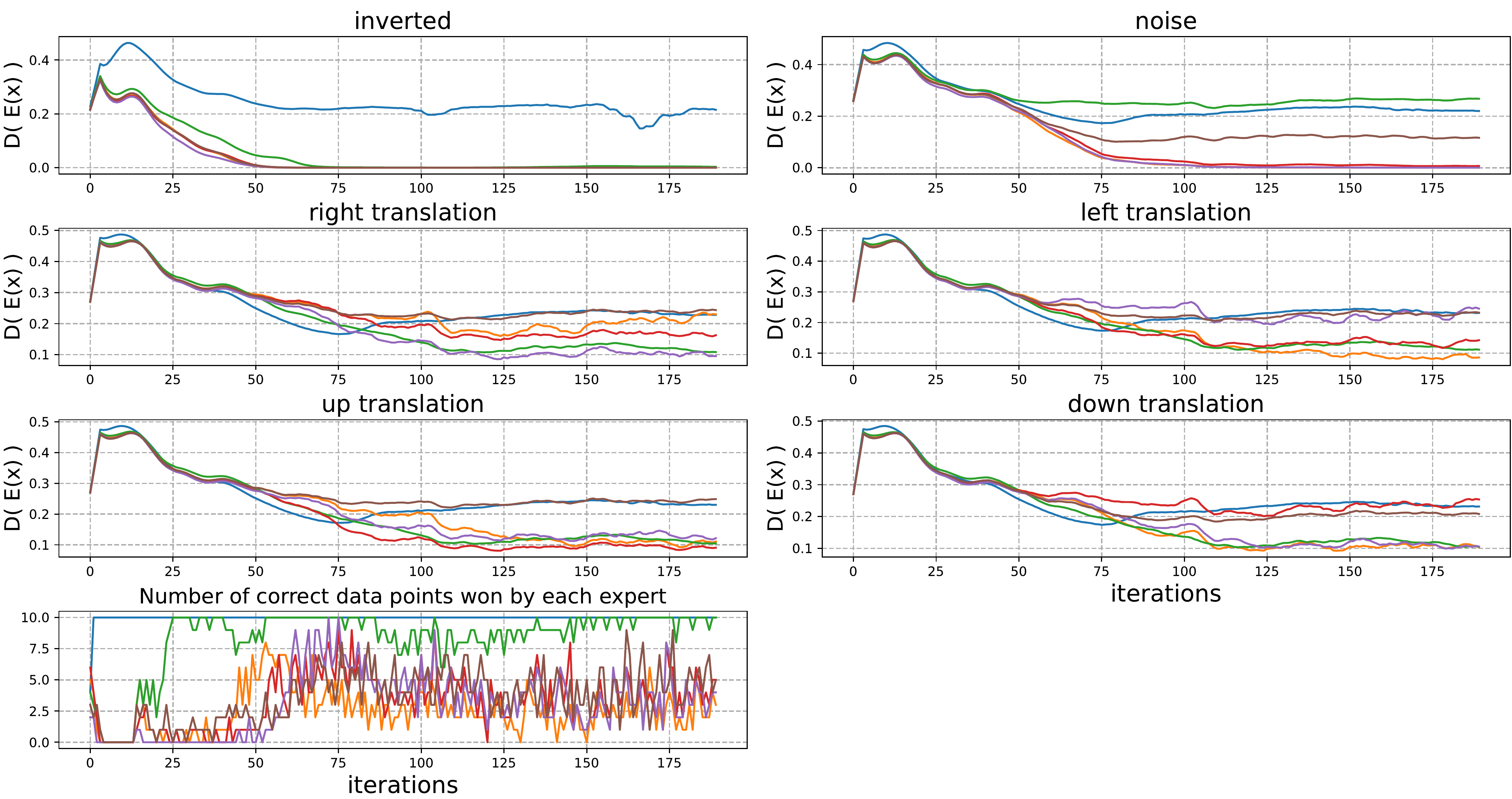}
  \caption{Without orthogonalization, transform severity 2px}
  \label{fig:severity_notorth_1px}
\end{figure}
\end{center}

\begin{figure}[t]
  \centering
  \includegraphics[width=\linewidth]{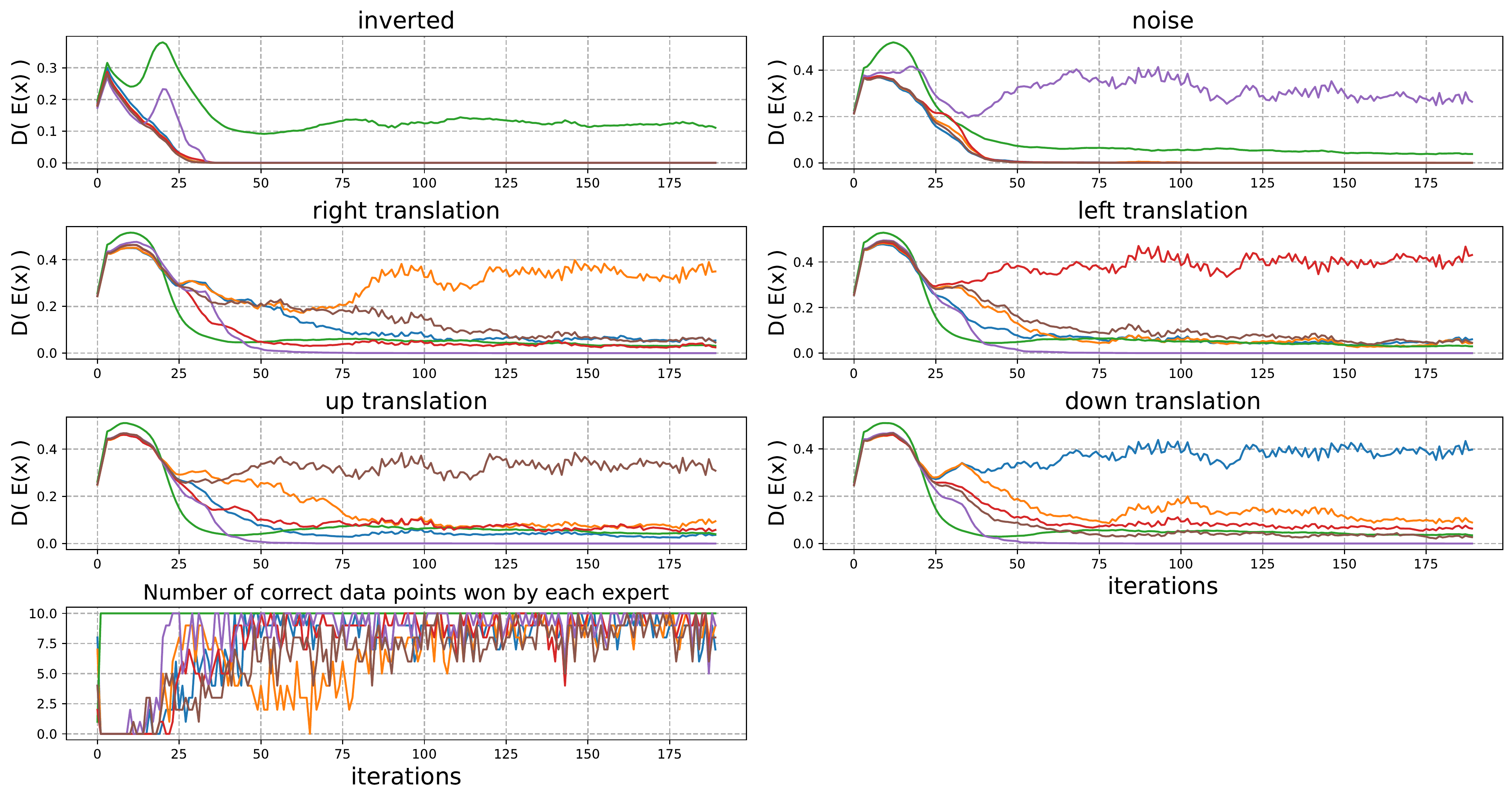}
  \caption{With orthogonalization, transform severity 2px}
  \label{fig:severity_orth_1px}
\end{figure}

Figures \ref{fig:severity_notorth_5px} and \ref{fig:severity_orth_5px} illustrate the results on severity of 5px, without and with orthogonalization, respectively. Furthermore, Figures \ref{fig:severity_notorth_1px} and \ref{fig:severity_orth_1px} illustrate corresponding results on severity of 2px.
For the more pronounced case, 5px, it can be seen that despite convergence speed being lower without orthogonalization, both methods converge nonetheless. On the other hand, in the case of 2px, the difference between the state of the art and the proposed method is no longer only terms of convergence time. Without orthogonalization, the experts are not able to converge at all, even given much higher number of iterations. It is clear that once the discriminator is unable to reliably identify the output of ``incorrect'' expert as such, a stronger diversity-promoting solution is necessary -- and orthogonalization can be just that.

\subsection{Data point relocation mechanism}

The final experiment is designed to justify the data point relocation mechanism as a way of discouraging the experts from claiming multiple transformations. It is worth noting that in these experiments, the relocation mechanism is accompanied by orthogonalization.

\begin{figure}[h!]
  \centering
  \includegraphics[width=\linewidth]{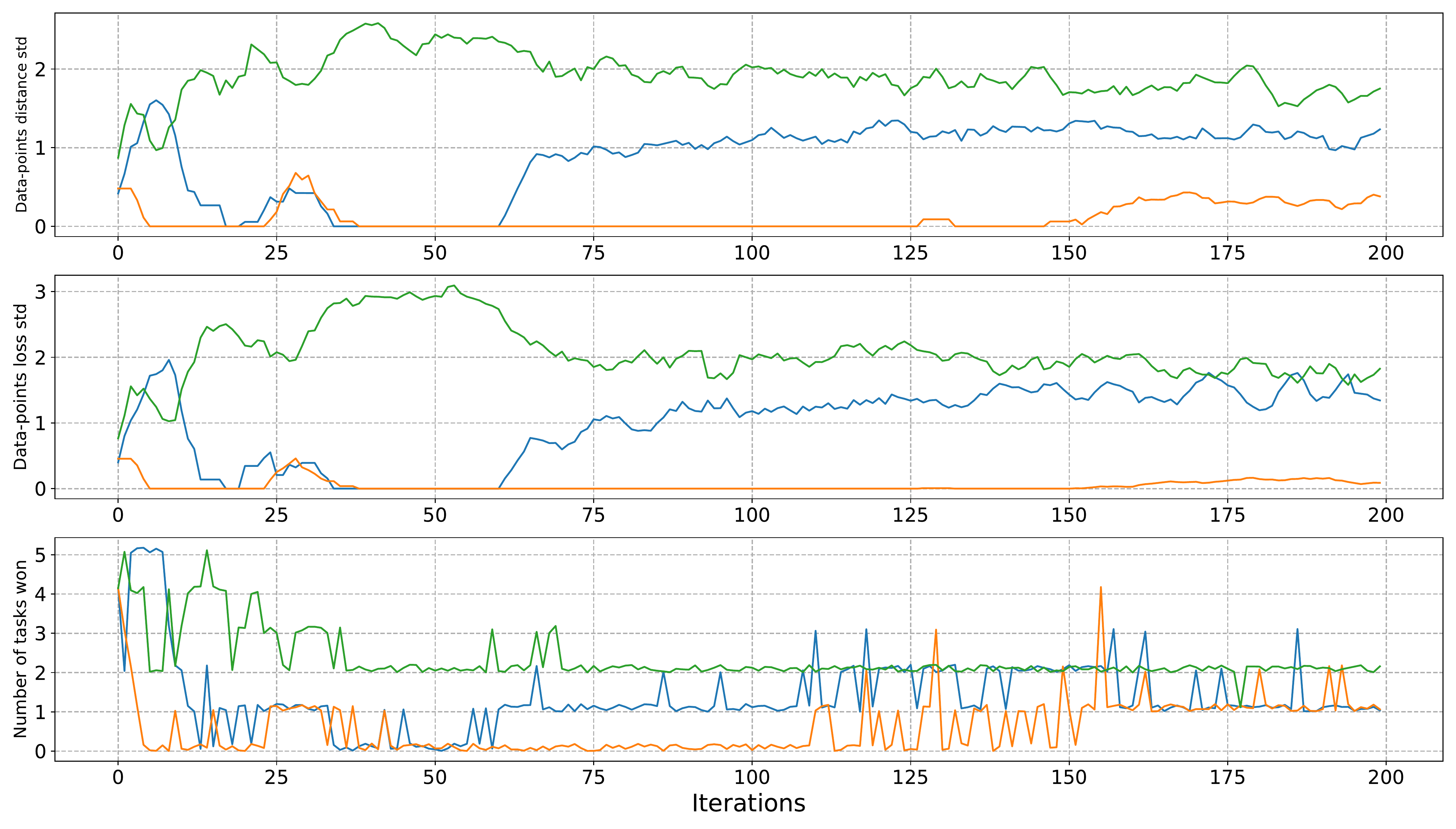}
  \caption{relation between number of transformations won by each expert and $std$ of last discriminator's hidden layer output (also $std$ of discriminator score) -- balanced sample}
  \label{fig:balancedsample}
\end{figure}

In the first part, we depict in Figure \ref{fig:balancedsample} the relationship between the standard deviation of the discriminator's last hidden layer output on the claimed data points (for different experts) and the number of transformations won by each expert. As an alternative that could be used instead, we also show the standard deviation of the discriminator score.
Note that, for the sake of readability, only three transformations, instead of all of them, are shown in the figure.

It is noticeable from Figure \ref{fig:balancedsample} that experts that win multiple transformations tend to have higher standard deviations. For instance, the expert associated with the blue curve, which starts by winning 4 or 5 transformations, shows very high standard deviation; while the expert shown in green, winning zero or one transformation, exhibits very low standard deviation. As the training progresses and experts converge towards winning one transformation each, their standard deviations get closer to each other.

In this experiment, the number of data points for each transformation within each batch of training is uniformly distributed. One might argue that once an expert starts winning more transformations, it will simply claim more data points; 
it could be the case that due to the sample prevalence of some transformations, the standard deviation becomes higher for experts winning those transformations, even if an expert wins one transformation. To show the robustness of the relocation mechanism against sample prevalence of transformations, in the next experiment for each training batch, the number of data points for one transformation is three times higher than the others. Figure \ref{fig:imbalancedsample} depicts three transformations one of which, left translation, has the largest number of data points. It can be seen that simply winning a larger number data points does not lead to increase in the standard deviation. Instead, as we hoped, the number of transformation won by the corresponding expert is the key factor affecting the standard deviation.

\subsection{Relocation percentage}

\begin{figure}[h]
  \centering
  \includegraphics[scale=0.17]{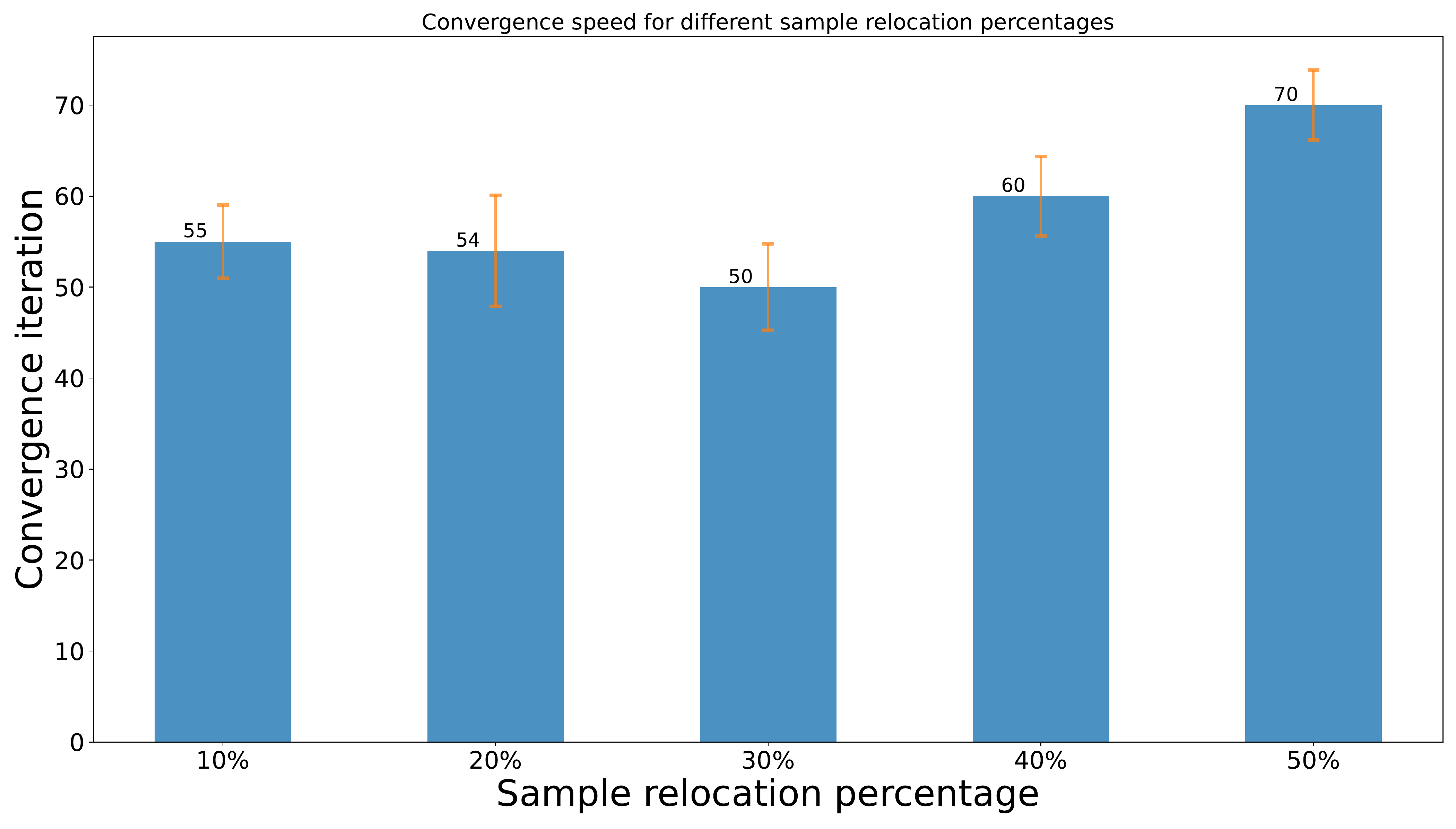}
  \caption{Effect of percentage of data points to relocate on convergence speed}
  \label{fig:relocation-percentage}
\end{figure}

Figure \ref{fig:relocation-percentage} depicts the sensitivity of the method on the percentage of data points relocated among experts ($rp$ in Algorithm \ref{alg:algorithm2}), in terms of convergence speed. Every bar is the average of five separate runs. It is noticeable that very high or very low relocation percentages deteriorate the convergence speed, however, selecting a reasonable value for this hyperparameter is not difficult. For the experiments in Section \ref{experiments}, $rp$ is selected to be $30\%$.

\subsection{Relocation mechanism robustness}

\begin{figure}[t]
  \centering
  \includegraphics[width=\linewidth]{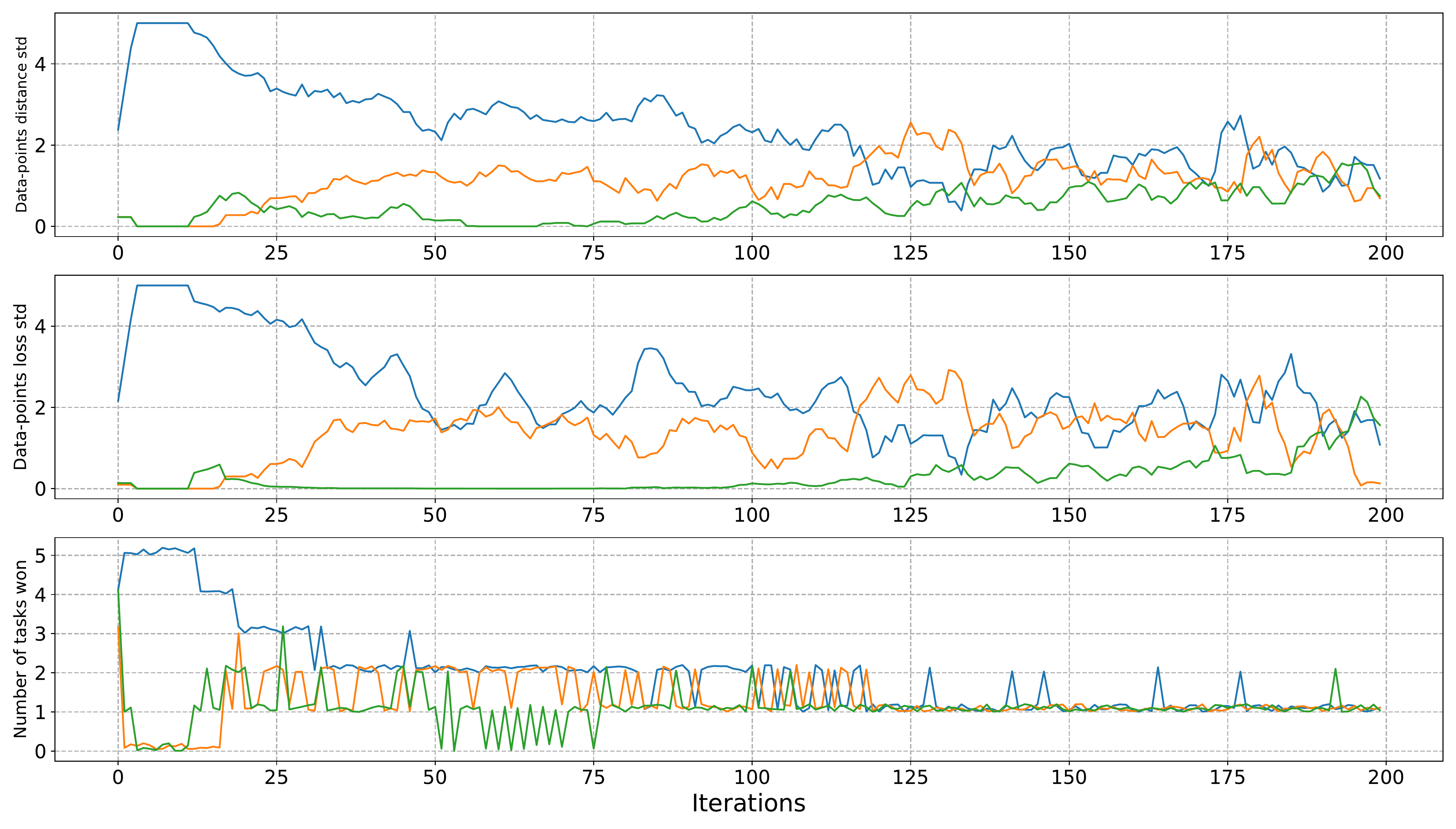}
  \caption{relation between number of transformations won by each expert and $std$ of last discriminator's hidden layer output (also $std$ of discriminator score) --  imbalanced sample}
  \label{fig:imbalancedsample}
\end{figure}

To show the robustness of the relocation mechanism against sample prevalence of transformations, in the next experiment for each training batch, the number of data points for one transformation is three times higher than the others. Figure \ref{fig:imbalancedsample} depicts three transformations one of which, left translation, has the largest number of data points. It can be seen that simply winning a larger number data points does not lead to increase in the standard deviation. Instead, as we hoped, the number of transformation won by the corresponding expert is the key factor affecting the standard deviation.

In this experiment, the number of data points for each transformation within each batch of training is uniformly distributed. One might argue that once an expert starts winning more transformations, it will simply claim more data points; 
it could be the case that due to the sample prevalence of some transformations, the standard deviation becomes higher for experts winning those transformations, even if an expert wins one transformation.  

\makeatletter
\setlength{\@fptop}{0pt}
\makeatother

\section{Conclusions}
\label{conclusion}

In this study, we have proposed a novel method for finding modular causal mechanisms. We have demonstrated that it converges significantly faster and has more power in identifying the modular mechanisms, compared to state of the art. 
We believe finding modular causal mechanisms is a key towards allowing AI systems to take the next step in generalization performance -- the kind of generalization that can go beyond the manifold of a single data set.

The primary constituent of our methodology is enforcing diversity between experts. Our key contribution is the observation that the performance bottleneck in this case lies at the insufficient power of the discriminator. A new method for reducing the similarity between experts was needed. This diversity is created by adopting an orthogonalization layer requiring the experts' outputs to be orthogonal to each other. To further reinforce modularization, we propose data points relocation, giving ``lost'' experts the opportunity to stay in the competition.

As a future work, our research naturally lends itself to be applied on more challenging data sets and in real-world scenarios. More importantly, the generalizability of the learned mechanism to new situations should be analyzed further. As a more technical future work direction, relaxing the orthogonalization constraint can also be considered. In other words, the level to which the experts' outputs needs to be de-correlated should possibly be adaptive, depending on the properties of the data and the transformations to be learned.

\begin{appendices}
\section{Detailed architectures and experts and discriminator}

Experts and discriminator architectures are shown in Tables \ref{tab:experts-arch} and \ref{tab:discriminator-arch}, respectively. The same architectures are used for both the model without orthogonalization and with orthogonalization. The difference is that with orthogonalization, as shown in Figure \ref{fig:podnnMechanism}, an orthogonalization layer is embedded after the fourth layer, to make experts outputs orthogonal to each other.

It is worth mentioning that we have also examined larger models for discriminator and/or for the experts, however, since the results (including the improvement of the proposed method over the baseline) is consistent across other architectures, we do not report this study in the paper in detail.

\begin{table}[h!]
\begin{center}
\caption{Architecture of the experts}
\begin{tabular}{ |c| c | c | c| c| } 
\hline
Layer type & Filters & Size/Stride  & Function & BN \\
\hline
Conv & 16 & 3x3/1 & ELU & \checkmark \\ 
\hline
Conv & 16 & 3x3/1 & ELU & \checkmark \\ 
\hline
Conv & 16 & 3x3/1 & ELU & \checkmark \\ 
\hline
Conv & 16 & 3x3/1 & ELU & \checkmark \\ 
\hline
Conv & 1 & 3x3/1 & Sigmoid &  \\ 
\hline
\end{tabular}
\end{center}

\label{tab:experts-arch}
\end{table}

\begin{table}[h]
\begin{center}
\caption{Discriminator architecture}
\begin{tabular}{ |c| c |  c | c| } 
\hline
Layer type &  \vtop{\hbox{\strut Filters(Conv)}\hbox{\strut Neurons(Dense)}}  & Size/Stride  & Function  \\
\hline
Conv & 16 & 3x3/1 & ELU  \\ 
Avg pooling & & 2x2/1 &   \\ 
\hline
Conv & 32 & 3x3/1 & ELU  \\ 
Avg pooling & & 2x2/1 &  \\  
\hline
Conv & 64 & 3x3/1 & ELU  \\ 
Avg pooling & & 2x2/1  &   \\ 
\hline
Dense & 128 & & ELU  \\ 
\hline
Dense & 1 &  & Sigmoid   \\ 
\hline
\end{tabular}
\end{center}

\label{tab:discriminator-arch}
\end{table}

The batch size during the training is selected to be 64. For all experiments, Adam optimizer is used \citep{kingma2018method}. For results presented in Figure \ref{fig:digits}, after the convergence of the experts, training is continued separately on the data points they claim for additional 100 iterations.

\section{Transformations details}

We used the following transformations for the experiments.

\begin{itemize}
\item Contrast inversion: The original value of each pixel, which is in the $[0,1]$ range, is transformed as ($1-$original value).
\item Noise addition: Gaussian noise with mean $0$ and standard deviation of $0.1$ is added.
\item Translations: eight directions (namely right, left, up, down, and their diagonals) are used. In Figure \ref{fig:convergence_speed_both}, 4px translation is used. In Figures \ref{fig:balancedsample} and \ref{fig:imbalancedsample}, 2px is used.
\end{itemize}
\end{appendices}

\bibliography{sn-article}

\begin{thebibliography}{}
\providecommand{\doi}[1]{\url{https://doi.org/#1}}
\bibcommenthead

\bibitem[\protect\citeauthoryear{Baldock, Maennel, and Neyshabur}{Baldock
  et~al.}{2021}]{baldock2021deep}
Baldock, R., H.~Maennel, and B.~Neyshabur. 2021.
\newblock Deep learning through the lens of example difficulty.
\newblock {\em Advances in Neural Information Processing Systems\/}~34:
  10876--10889 .

\bibitem[\protect\citeauthoryear{Battaglia, Hamrick, Bapst, Sanchez-Gonzalez,
  Zambaldi, Malinowski, Tacchetti, Raposo, Santoro, Faulkner, et~al.}{Battaglia
  et~al.}{2018}]{battaglia2018relational}
Battaglia, P.W., J.B. Hamrick, V.~Bapst, A.~Sanchez-Gonzalez, V.~Zambaldi,
  M.~Malinowski, A.~Tacchetti, D.~Raposo, A.~Santoro, R.~Faulkner, et~al. 2018.
\newblock Relational inductive biases, deep learning, and graph networks.
\newblock {\em arXiv preprint arXiv:1806.01261\/} .

\bibitem[\protect\citeauthoryear{Besserve, Mehrjou, Sun, and
  Sch{\"o}lkopf}{Besserve et~al.}{2018}]{besserve2018counterfactuals}
Besserve, M., A.~Mehrjou, R.~Sun, and B.~Sch{\"o}lkopf. 2018.
\newblock Counterfactuals uncover the modular structure of deep generative
  models.
\newblock {\em arXiv preprint arXiv:1812.03253\/} .

\bibitem[\protect\citeauthoryear{Chollet}{Chollet}{2019}]{chollet2019measure}
Chollet, F. 2019.
\newblock On the measure of intelligence.
\newblock {\em arXiv preprint arXiv:1911.01547\/} .

\bibitem[\protect\citeauthoryear{Dhillon and Verma}{Dhillon and
  Verma}{2020}]{dhillon2020convolutional}
Dhillon, A. and G.K. Verma. 2020.
\newblock Convolutional neural network: a review of models, methodologies and
  applications to object detection.
\newblock {\em Progress in Artificial Intelligence\/}~{\em 9\/}(2): 85--112 .

\bibitem[\protect\citeauthoryear{Fox and Wiens}{Fox and
  Wiens}{2019}]{fox2019advocacy}
Fox, I. and J.~Wiens. 2019.
\newblock Advocacy learning: Learning through competition and class-conditional
  representations.
\newblock {\em arXiv preprint arXiv:1908.02723\/} .

\bibitem[\protect\citeauthoryear{Goodfellow, Lee, Le, Saxe, and Ng}{Goodfellow
  et~al.}{2009}]{goodfellow2009measuring}
Goodfellow, I., H.~Lee, Q.~Le, A.~Saxe, and A.~Ng. 2009.
\newblock Measuring invariances in deep networks.
\newblock {\em Advances in neural information processing systems\/}~22:
  646--654 .

\bibitem[\protect\citeauthoryear{Goodfellow, Pouget-Abadie, Mirza, Xu,
  Warde-Farley, Ozair, Courville, and Bengio}{Goodfellow
  et~al.}{2020}]{goodfellow2020generative}
Goodfellow, I., J.~Pouget-Abadie, M.~Mirza, B.~Xu, D.~Warde-Farley, S.~Ozair,
  A.~Courville, and Y.~Bengio. 2020.
\newblock Generative adversarial networks.
\newblock {\em Communications of the ACM\/}~{\em 63\/}(11): 139--144 .

\bibitem[\protect\citeauthoryear{Jakubovitz, Giryes, and Rodrigues}{Jakubovitz
  et~al.}{2019}]{jakubovitz2019generalization}
Jakubovitz, D., R.~Giryes, and M.R. Rodrigues. 2019.
\newblock Generalization error in deep learning, {\em Compressed Sensing and
  Its Applications},  153--193. Springer.

\bibitem[\protect\citeauthoryear{Kingma and Adam}{Kingma and
  Adam}{2018}]{kingma2018method}
Kingma, D.P. and B.J. Adam. 2018.
\newblock a method for stochastic optimization. 2014.
\newblock {\em arXiv preprint arXiv:1412.6980\/}~9 .

\bibitem[\protect\citeauthoryear{Locatello, Vincent, Tolstikhin, R{\"a}tsch,
  Gelly, and Sch{\"o}lkopf}{Locatello et~al.}{2018}]{locatello2018competitive}
Locatello, F., D.~Vincent, I.~Tolstikhin, G.~R{\"a}tsch, S.~Gelly, and
  B.~Sch{\"o}lkopf. 2018.
\newblock Competitive training of mixtures of independent deep generative
  models.
\newblock {\em arXiv preprint arXiv:1804.11130\/} .

\bibitem[\protect\citeauthoryear{Mashhadi, Nowaczyk, and Pashami}{Mashhadi
  et~al.}{2021}]{mashhadi2021parallel}
Mashhadi, P.S., S.~Nowaczyk, and S.~Pashami. 2021.
\newblock Parallel orthogonal deep neural network.
\newblock {\em Neural Networks\/}~140: 167--183 .

\bibitem[\protect\citeauthoryear{Nowaczyk, R{\"o}gnvaldsson, Fan, and
  Calikus}{Nowaczyk et~al.}{2020}]{Nowaczyk2020}
Nowaczyk, S., T.~R{\"o}gnvaldsson, Y.~Fan, and E.~Calikus 2020.
\newblock {\em Towards Autonomous Knowledge Creation from Big Data in Smart
  Cities}, pp.\  1--35.
\newblock Cham: Springer International Publishing.

\bibitem[\protect\citeauthoryear{Parascandolo, Kilbertus, Rojas-Carulla, and
  Sch{\"o}lkopf}{Parascandolo et~al.}{2018}]{parascandolo2018learning}
Parascandolo, G., N.~Kilbertus, M.~Rojas-Carulla, and B.~Sch{\"o}lkopf 2018.
\newblock Learning independent causal mechanisms.
\newblock In {\em International Conference on Machine Learning}, pp.\
  4036--4044. PMLR.

\bibitem[\protect\citeauthoryear{Sch{\"o}lkopf}{Sch{\"o}lkopf}{2019}]{scholkopf2019causality}
Sch{\"o}lkopf, B. 2019.
\newblock Causality for machine learning.
\newblock {\em arXiv preprint arXiv:1911.10500\/} .

\bibitem[\protect\citeauthoryear{Vankov and Bowers}{Vankov and
  Bowers}{2020}]{vankov2020training}
Vankov, I.I. and J.S. Bowers. 2020.
\newblock Training neural networks to encode symbols enables combinatorial
  generalization.
\newblock {\em Philosophical Transactions of the Royal Society B\/}~{\em
  375\/}(1791): 20190309 .

\bibitem[\protect\citeauthoryear{von K{\"u}gelgen, Ustyuzhaninov, Gehler,
  Bethge, and Sch{\"o}lkopf}{von K{\"u}gelgen et~al.}{2020}]{von2020towards}
von K{\"u}gelgen, J., I.~Ustyuzhaninov, P.~Gehler, M.~Bethge, and
  B.~Sch{\"o}lkopf. 2020.
\newblock Towards causal generative scene models via competition of experts.
\newblock {\em arXiv preprint arXiv:2004.12906\/} .

\bibitem[\protect\citeauthoryear{Xie, Ras, van Gerven, and Doran}{Xie
  et~al.}{2020}]{xie2020explainable}
Xie, N., G.~Ras, M.~van Gerven, and D.~Doran. 2020.
\newblock Explainable deep learning: A field guide for the uninitiated.
\newblock {\em arXiv preprint arXiv:2004.14545\/} .

\end{thebibliography}


\end{document}